\documentclass{article}

\usepackage[preprint]{neurips_2022}

\usepackage[utf8]{inputenc} 
\usepackage[T1]{fontenc}    
\usepackage{hyperref}       
\usepackage{url}            
\usepackage{booktabs}       
\usepackage{amsfonts}       
\usepackage{nicefrac}       
\usepackage{microtype}      
\usepackage{xcolor}         
\usepackage{amsmath}         
\usepackage{multirow}
\usepackage{mathtools}
\usepackage{epsfig} 
\usepackage{caption}
\usepackage{subcaption}
\usepackage{comment}

\newcommand\methodA{GMPool}
\newcommand\methodB{NGMPool}

\title{Grouping-matrix based Graph Pooling \\with Adaptive Number of Clusters}

\author{
  Sung Moon Ko$^1$ \quad Sungjun Cho$^1$ \quad Dae-Woong Jeong$^1$\\
  \textbf{Sehui Han}$^1$ \quad \textbf{Moontae Lee}$^{1,2}$ \quad \textbf{Honglak Lee}$^{1}$\\
  $^1$LG AI Research \quad $^2$University of Illinois Chicago\\
}

\begin{document}

\maketitle
\begin{abstract}
Graph pooling is a crucial operation for encoding hierarchical structures within graphs. Most existing graph pooling approaches formulate the problem as a node clustering task which effectively captures the graph topology. Conventional methods ask users to specify an appropriate number of clusters as a hyperparameter, then assume that all input graphs share the same number of clusters. In inductive settings where the number of clusters can vary, however, the model should be able to represent this variation in its pooling layers in order to learn suitable clusters. Thus we propose \textsc{GMPool}, a novel differentiable graph pooling architecture that automatically determines the appropriate number of clusters based on the input data. The main intuition involves a {\it grouping matrix} defined as a quadratic form of the pooling operator, which induces use of binary classification probabilities of pairwise combinations of nodes. \textsc{GMPool} obtains the pooling operator by first computing the grouping matrix, then decomposing it. Extensive evaluations on molecular property prediction tasks demonstrate that our method outperforms conventional methods.
\end{abstract}

\section{Introduction}
Graph Neural Networks (GNNs) learn representations of individual nodes based on the connectivity structure of an input graph. 
For graph-level prediction tasks, the standard procedure globally pools all the node features into a single graph representation without weight difference, then feeds the representation to a final prediction layer. This process implies that information only propagates through node-to-node edges, rendering the model unable to hierarchically aggregate information efficiently beyond local convolution.

However, a hierarchical structure can encode the global topology of graphs that is useful for effective learning of long range interactions. Therefore, designing a pooling architecture which respects the graph structure is crucial for downstream tasks such as social network analyses \cite{https://doi.org/10.48550/arxiv.1609.02907, https://doi.org/10.48550/arxiv.1706.02216} and molecule property predictions \cite{https://doi.org/10.48550/arxiv.1509.09292, https://doi.org/10.48550/arxiv.1606.09375, https://doi.org/10.48550/arxiv.1312.6203, doi:10.1021/acs.jcim.6b00601, 4700287, https://doi.org/10.48550/arxiv.1812.01070}. 

As an alternative to global pooling, DiffPool first proposed an end-to-end differentiable pooling by soft-classifying each node into a smaller number of clusters \cite{ying2018}. Later gPool \cite{gao2019} and SAGpool \cite{lee2019} incorporated the attention mechanism into pooling, while MinCutPool proposed grouping the nodes into clusters by minimizing the relaxed $K$-way normalized minimum cut objective \cite{bianchi2019}.

In most inductive settings, there is no single number of clusters that is suitable across all graphs in the dataset.
Particularly in molecular graphs, the number of functional groups often determines useful characteristics and chemical behaviors, while varying significantly across different molecules.
Nonetheless, existing pooling methods require the number of clusters as a hyperparameter, then operates under the assumption that all graphs share the same number of clusters~\cite{ranjan2020asap}. This is often undesirable as it not only requires additional hyperparameter tuning, but also imposes a strong inductive bias that deteriorates downstream performance.

To overcome this challenge, we propose \methodA{}, a general pooling framework that does not require an universal number of clusters as a user hyperparameter. Figure 1 depicts the overall framework of \methodA{}. The core intuition is that the product of a pooling matrix with itself forms a grouping matrix, where each $(i,j)$-th entry indicates the pairwise {\it clustering similarity}: whether the nodes $i$ and $j$ are pooled to the same clusters. For each graph, \methodA{} parameterizes the clustering similarities in its grouping matrix via a classification layer. Finally, we perform SVD on the grouping matrix to obtain the pooling matrix such that the overall rank represents the suitable number of clusters. We also test a single-pooling variant \methodB{} that does not perform any decomposition, but rather uses the grouping matrix as is. In real-world molecular property prediction tasks, we show that our approach outperforms previous baselines, while successfully learning suitable clusters.

\begin{figure}[!t]
    \centering
    \includegraphics[height=40mm, width=130mm, scale=0.5, trim={3.8cm 9cm 2cm 2cm}, clip]{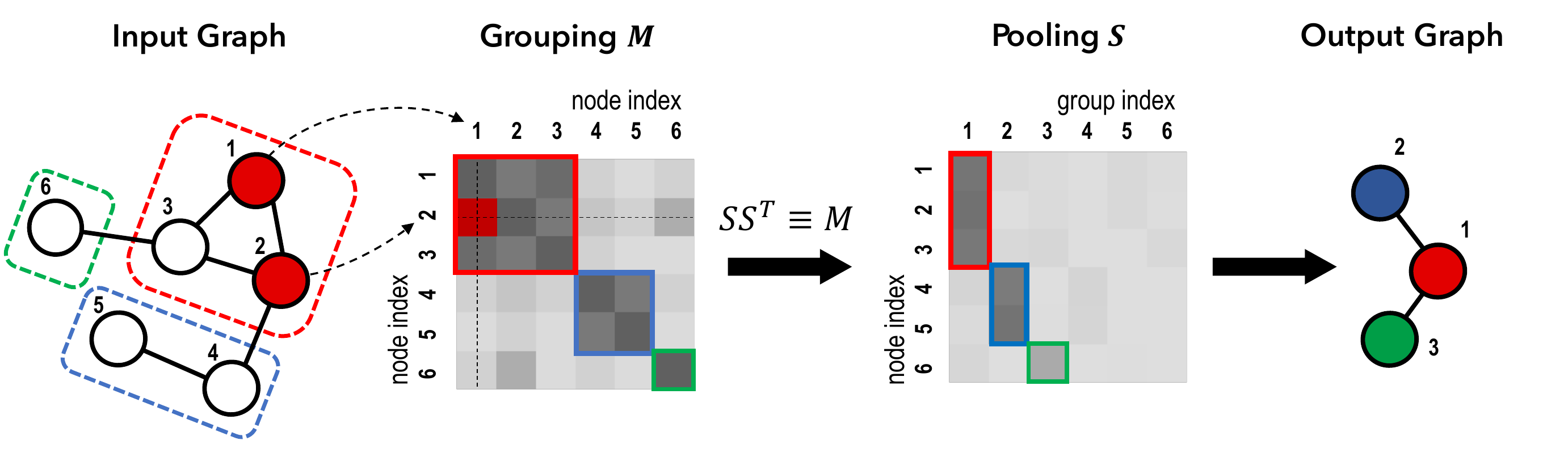}
    \caption{An illustration of our framework. Our method first forms a grouping matrix that encodes clustering similarities between each pair of nodes and then acquires the pooling matrix that coarsens the graph by decomposing the grouping matrix.}
    \label{fig:algorithm}
\end{figure}

The main contributions of this paper are as follows:
\begin{itemize}
    \item We design a grouping matrix-based pooling operator that does not require users to specify the number of clusters a priori.
    \item We propose \methodA{} and \methodB{}. \methodA{} performs SVD on the grouping matrix to obtain the pooling matrix, whereas \methodB{} utilizes the grouping matrix as is.
    \item We demonstrate the power of our methods both quantitatively and qualitatively on a wide range of real molecular property prediction tasks.
\end{itemize}

\section{Related Work}
\paragraph{Graph Neural Networks.} GNN architectures have shown great performance in various fields such as social network data, authorship/citation networks, and molecular data that can naturally be interpreted as graphs. For graph convolution, several work have utilized the graph Laplacian in the spectral domain. However, sheer convolution in the spectral domain suffers from the non-locality problem, and various approaches have been introduced to overcome this limitation. \cite{https://doi.org/10.48550/arxiv.1706.02216, https://doi.org/10.48550/arxiv.1810.00826, https://doi.org/10.48550/arxiv.1710.10903, https://doi.org/10.48550/arxiv.1704.01212} One stream of work has embedded the attention architecture into GNN, inferring the interaction between nodes without using a diffusion-like picture. \cite{https://doi.org/10.48550/arxiv.1710.10903} Another line of work considered message passing networks, which ensures the signal to be localized and non-linearly weighted. \cite{https://doi.org/10.48550/arxiv.1704.01212} This architecture has been proven to be highly effective in molecular property prediction fields. \cite{doi:10.1021/acs.jcim.9b00237} \

\paragraph{Graph Pooling.} Graph pooling aims to utilize the hierarchical nature of graphs. Early work mainly focused on fixed axiomatic pooling methods such as minimum cut, k-means, and spectral clustering without any gradient-based optimization. \cite{https://doi.org/10.48550/arxiv.1312.6203, NIPS2011_6c1da886, 10.1016/j.patcog.2006.04.007, https://doi.org/10.48550/arxiv.0711.0189, 4302760} 
Although these pooling methods are effective on graphs without noise, the same heuristic often fails to work well on real datasets and tasks, especially due to a lack of differentiability that prohibits training under supervised signals.
Since node representations and pooling strategies mutually affect each other during the training process, simultaneous optimization of whole components is crucial for avoiding local minima. Among many solutions, Diffpool\cite{ying2018} is the first to propose an end-to-end learnable pooling mechanism that learns an assignment matrix in which each entry represents the probability of a node being assigned to a cluster.

gPool \cite{gao2019} and SAGPool \cite{lee2019} are ranking-based pooling methods that coarsen the input graph by ranking and downsampling a small subset of nodes. MinCutPool \cite{bianchi2019} leverages a continuous relaxation of the minimum-cut objective, enabling spectral clustering under full differentiability.

However, the pooling methods above all share a common limitation: the number of clusters must be predefined for each layer as hyperparameters. This limitation is especially detrimental in inductive settings such as molecular property prediction, where each graph can have varying numbers of useful sub-structures. \cite{https://doi.org/10.1111/cbdd.12952, doi:10.1021/acs.jmedchem.0c00754, GUVENCH20161928} Allowing the model to pool towards varying number of clusters based on data is expected to enhance performance, and our proposed \methodA{} allows such variation through the rank of the grouping matrix. To the best of our knowledge, \methodA{} is the first to achieve high performance without the need to manually adjust the number of clusters through additional hyperparameter tuning.

\section{Proposed Method}
\label{method}
In this section, we propose a novel differentiable pooling layer, \methodA{}, which obtains the pooling matrix by first building a grouping matrix that contains clustering similarities of pairwise nodes and then decomposing the matrix into its square-root form. We start the section with preliminary information, then outline the details of \methodA{} in later sections.

\subsection{Preliminaries}\label{sec:preliminaries}

\paragraph{Problem setting.} We assume an inductive graph-level prediction setting where our aim is to learn a function $f_\theta: \mathcal{G} \to \mathcal{Y}$ that maps a graph $G \in \mathcal{G}$ to a property label $y \in \mathcal{Y}$.
Each graph $G$ with $n$ nodes is represented as a triplet $G = (A, X, E)$ with graph adjacency $A \in \{0,1\}^{n\times n}$, node features $X \in \mathbb{R}^{n \times d_n}$, and edge features $E \in \mathbb{R}^{n \times n \times d_e}$. We use $X_i$ and $E_{ij}$ to denote the features of node $i$ and edge $(i,j)$, respectively. 

\paragraph{Directed MPNN and pooling.} As our backbone GNN, we adopt the Directed Message Passing Neural Network (DMPNN) \cite{doi:10.1021/acs.jcim.9b00237} which aggregates messages through directed edges. Note that while we chose DMPNN due to its superior performance over GNN architectures, our pooling layer is module-agnostic and can be combined with any GNN as long as node representations are returned as output. 
Given a graph, DMPNN first initializes the hidden state of each edge $(i,j)$ based on its feature $E_{ij}$ and the source-node's feature $X_i$. At each timestep $t$, each directional edge gathers hidden states from incident edges into a message $m_{ij}^{t+1}$ and updates its own hidden state to $h_{ij}^{t+1}$ as follows
\begin{gather}
    m_{ij}^{t+1} = \sum_{k \in \mathcal{N}(i)\setminus j} h_{ki}^t\\
    h_{ij}^{t+1} = \texttt{ReLU}(h_{ij}^0 + W_e m_{ij}^{t+1})
\end{gather}
Here, $\mathcal{N}(i)$ denotes the set of neighboring nodes of node $i$ and $W_e$ a learnable weight. The hidden states of nodes are updated by aggregating the hidden states of incident edges into message $m_i^{t+1}$, and passing its concatenation with the node feature $X_i$ into a linear layer followed by ReLU non-linearity
\begin{gather}
    m_i^{t+1} = \sum_{j \in \mathcal{N}(i)} h_{ij}^t\\
    h_i^{t+1} = \texttt{ReLU}(W_n \texttt{concat}(X_i, m_i^{t+1})) 
\end{gather}
Similarly, $W_n$ denotes a learnable weight. Assuming DMPNN runs for $T$ timesteps, we use $(X_{out},E_{out}) = \texttt{GNN}(A, X, E)$ to denote the output representation matrices containing hidden states of all nodes and edges, respectively (i.e., $X_{out,i} = h_i^T$ and $E_{out,ij} = h_{ij}^T$).

For graph-level prediction, the node representations after the final GNN layer are typically sum-pooled to obtain a single graph representation $h_G = \sum_{i} h_i$, which is then passed to a FFN prediction layer. Note that this approach only allows features to propagate locally and is hence unable to learn long-range dependencies and hierarchical structures within graphs.

Our goal is to learn a pooling operator to coarsen the input graph after the GNN in each hierarchical layer. In each hierarchical layer, the GNN constructs node representations and then the pooling layer forms a coarsened graph, which is used as input to the next hierarchical layer. More formally, given the representations from the $l$-th layer as $(X^{(l)}_{out}, E^{(l)}_{out}) = \texttt{GNN}(A^{(l)}, X^{(l)}, E^{(l)})$, the pooling layer yields an assignment matrix $S^{(l)} \in \mathbb{R}^{n_l \times n_{l+1}}$ pooling $n_l$ nodes into $n_{l+1}$ clusters. Then, the graph $G^{(l)} = (A^{(l)}, X^{(l)}, E^{(l)})$ is coarsened into $G^{(l+1)} = (A^{(l+1)}, X^{(l+1)}, E^{(l+1)}) = (S^{(l)^T} A^{(l)} S^{(l)}, S^{(l)^T} X^{(l)}_{out}, S^{(l)^T} E^{(l)}_{out} S^{(l)})$. This hierarchical process can be utilized iteratively depending on the task at hand.

\subsection{Differentiable Pooling via Grouping Decomposition}
When looking into the relation between pairs of nodes, the grouping task becomes rather simple. While most previous models focus on classifying each node to a predefined number of clusters, our idea simplifies the task into classifying whether each pair of nodes is in the same group. Thus, setting the number of clusters a priori becomes unnecessary. This classification will go through every pair of combinations of nodes to ensure permutation invariance.  
\begin{equation}
\label{gm}
    M^{(l)}_{ij} = \textrm{Softmax}(\textrm{Clf}(f(X_i, X_j))) \qquad \forall \,\, i, j \in \textrm{\# of Nodes}
\end{equation}
where $M^{(l)} \in \mathbb{R}^{N \times N}$ and $f$ is a commutative function
\begin{equation}
    f : X \oplus X \rightarrow Y \qquad \textrm{where} \,\, X, Y \in \mathbb{R}^N
\end{equation}
that maps two input vectors into one output vector. While there exist many available choices for $f$, we use Euclidean distance between input vectors to simplify the classification task. Each matrix index corresponds to the node number and each element contains probability values for each pair of nodes whether they are in the same group. 

As an illustrative example, consider a set of disjoint clusters with no overlapping nodes. In such case, the grouping matrix not only contains $0, 1$ as its elements, but also can be reformed into a block diagonal form. The number of blocks corresponds to the number of groups after pooling and nodes assigned to the same blocks corresponds to a same group. For instance, if there are three different groups and each group size are $k_1, k_2, k_3$,
\begin{equation}
  \setlength{\arraycolsep}{0pt}
  M^{(l)} = \begin{bmatrix}
    \fbox{$1_{k_1 \times k_1}$} & 0 & 0  \\
    0 & \fbox{$1_{k_2 \times k_2}$} & 0 \\
    0 & 0 & \fbox{$1_{k_3 \times k_3}$} \\
  \end{bmatrix}
  \label{groupingmatrix}
\end{equation}
One can easily see that the corresponding pooling operator is as follows
\begin{equation}
  \setlength{\arraycolsep}{0pt}
  S^{(l)} = \begin{bmatrix}
    \fbox{$1_{k_1 \times 1}$} & 0 & 0 \quad & \cdots & \quad 0\\
    0 & \fbox{$1_{k_2 \times 1}$} & 0 \quad & \cdots & \quad 0\\
    0 & 0 & \fbox{$1_{k_3 \times 1}$} \quad & \cdots & \quad 0\\
  \end{bmatrix}
\end{equation}
In general, each element of the grouping matrix (in eq. \ref{groupingmatrix}) is a continuous number within $[0, 1]$, which allows soft-clustering with overlapping nodes. For detailed computation, see appendix.

However, the grouping matrix itself has a limited role in pooling operations. Therefore, extracting pooling operators from the grouping matrix is crucial. Our strategy to form a pooling operator is rather simple. It can be acquired by decomposing a grouping matrix into square-root form. There are numerous known methods which can be utilized, yet we will introduce two representative methods in the following subsection.


\subsection{Decomposition Schemes}
While the grouping matrix cannot be used for pooling as is, it encodes how similarly each pair of nodes are pooled as it equals the product of the pooling operator with its transpose.
The $(i,j)$-th entry of the grouping matrix equals $\langle S_i^{(l)}, S_j^{(l)}\rangle = 1$ if the nodes are exactly pooled to the same clusters, $\langle S_i^{(l)}, S_j^{(l)} \rangle = 0$ if they are pooled orthogonally to different clusters. Therefore, if we can decompose the grouping matrix into square-root form, it can be interpreted as a pooling operator for the model.
\begin{equation}
   \label{square} S^{(l)}S^{(l)T} = M^{(l)}
\end{equation}
The pooling operator $S \in \mathbb{R}^{n_{l} \times n_{l+1}}$ is a matrix where $n_{l+1} \le n_{l}$. Note that by multiplying pooling operator $S$ in reverse order, the degree matrix $D \in \mathbb{R}^{n_{l+1}\times n_{l+1}}$ of pooling space can be obtained.
\begin{equation}
    \label{degree} S^{(l)T}S^{(l)} = D^{(l)}
\end{equation}
From eq. \ref{square}, it is obvious that the pooling operator completely reconstructs grouping matrix by interacting pooling indices. Moreover, $S$ can be interpreted as a weighted matrix for each node to form appropriate sub-structures.

\begin{table*}
  \caption{Dataset Summary}
  \label{data_sum}
  \centering
  \begin{tabular}{llllll}
    \toprule
    Dataset     & Description     & Input Type    & Mean Size    & Task Type    & Data \#\\
    \midrule
    PLQY & Fluorescent   & SMILES    & 57.3    & Regression    & 515\\
    $\lambda_{max}$ Solvents     & Fluorescent  & SMILES    & 55.2    & Regression    & 1,070\\
    $\lambda_{max}$ Films     & Fluorescent        & SMILES    & 57.2    & Regression     & 1,270\\
    pIC50     & Binding       & SMILES    & 30.5    & Regression    & 10,978\\
    Tox21     & Toxicity       & SMILES    & 26.0    & Classification    & 2,514\\
    \bottomrule
  \end{tabular}
\end{table*}

\subsubsection{Eigen Decomposition Based Method}
\label{eigenmethod}
Eigen decomposition is one of the basic decomposition schemes one can consider. It is widely used to decompose a given matrix into orthonormal basis $O \in \mathbb{R}^{n_l \times n_l}$ and eigen value $\Lambda \in \mathbb{R}^{n_l \times n_l}$.
\begin{equation} \label{eigen}
    M^{(l)} = O \Lambda O^T
\end{equation}
This particular decomposition scheme always works unless the determinant of a given matrix is equal to 0. From eq. \ref{eigen}, one can rearrange RHS of the equation to become a square form of pooling operator if we set $n_{l + 1} = n_l$.
\begin{equation}
    M^{(l)} = O \sqrt{\Lambda} \sqrt{\Lambda} O^T\equiv S^{(l)} S^{(l)T}
\end{equation}
The pooling operator $S$ is a square matrix with size of $n_l \times n_l$, yet the eigen value $\Lambda$ suppresses useless ranks in the matrix by multiplying $0$ to each column of orthonormal basis. 
Also, eigen decomposition works for any matrix with non-zero determinants, and so it performs perfectly fine in real world situations. Furthermore, any symmetric and real matrix are guaranteed to have real eigen values as well as vectors. Therefore, the square-root of the grouping matrix is ensured to be interpreted as a transformation operator forming sub-groups from nodes. These continuous real valued elements have the advantage that nodes can be soft-clustered to sub-groups.
In conventional clustering, it is hard to cluster these structures properly. However, since soft clustering is naturally embedded in the algorithm, linker structures can be dealt with ease.    

After acquiring the pooling operator, the pooling process becomes obvious. Nodes are in fundamental representation while edge features and adjacency matrix are in adjoint representation. Which leads to the following transformation rules.
\begin{gather}
    X_i^{(l+1)} = S^{(l)} X_i^{(l)}\\
    E_{ij}^{(l+1)} = S^{(l)} E_{ij}^{(l)} S^{(l)T}\\
    A_{ij}^{(l+1)} = S^{(l)} A_{ij}^{(l)} S^{(l)T}
\end{gather}
If grouping is properly done, $0$ (or close to $0$) components will appear in the decomposed eigen value matrix. These zero eigenvalues arise naturally and play a role in disregarding group information; those are ineffective towards prediction. However, zero elements in the eigen values causes a major problem in the decomposition process since the matrix might carry a singular determinant.
Eigen decomposition is based on an iterative approximation algorithm which includes unbounded terms if any two eigen values are small or close. One can see clearly about this matter in \cite{DBLP:journals/corr/IonescuVS15}.
\begin{equation}
    \Big(\frac{\partial{l}}{\partial{A}}\Big) = U\big(K^T \odot (U^T \frac{\partial{l}}{\partial{U}}) + (\frac{\partial{l}}{\partial{\Lambda}})_{\textrm{diag}}) (U^T)
\end{equation}
Here, $\odot$ denotes element-wise product. Off-diagonal components of $K = 1/(\lambda_i - \lambda_j)$ causes the problem, since the value blows up to the infinity if any two eigen values are close or very small. However, there are some solutions for this matter by approximating gradient in different ways \cite{DBLP:journals/corr/abs-1906-09023, 9400752, DBLP:journals/corr/abs-2105-02498}. Those methods are developed further to achieve higher speed in the calculation \cite{DBLP:journals/corr/abs-2201-08663}. They claim that the method is noticeably faster, over $8$ times, than the standard SVD which has the time complexity $\mathcal{O}(n^3)$. Thus, we utilized this method in our work to stabilize and accelerate the learning process. However, since the algorithm achieves the higher speed by approximating gradients, the error compared to standard SVD grows bigger as the size of the matrix grows. Therefore, this method might not be valid with large sized graph data.


\subsubsection{Pooling without Decomposing The Grouping Matrix}
Another decomposition scheme we are introducing has a rather different approach. Since computing the square root of a given matrix is not an easy task, here we focus on the square of the pooling operator, which is nothing but the grouping matrix itself, and formulate a pooling-like effect by multiplying the grouping matrix. The key idea is to retain pooling depth to one and use a weighted aggregation vector in pooling space as an aggregation basis. The weighted aggregation vector is transformed Euclidean one vector by acting a pooling matrix obtained by decomposing the grouping matrix.
\begin{gather}
    1_i^{(l+1)} = S^{(l)}1_i^{(l)}
\end{gather}
The final form of the transformation can be expressed as follows. 
\begin{gather}
    X_i^{(l+1)} \sim M^{(l)} X_i^{(l)} \label{ngmp-node}\\
    E_{ij}^{(l+1)} \sim M^{(l)} E_{ij}^{(l)} M^{(l)} \label{ngmp-edge}
\end{gather}
 This pooling scheme is simpler to use and more scalable (with $\mathcal{O}(n^2)$ cost) than GMPool since the method circumvents SVD computation. Yet there are two mathematical ambiguities. One is that it is only valid for single depth pooling cases. If one tries to perform multiple sequential pooling operations, the pooling operators are no more available to be reduced into the grouping matrix, since two different pooling operators are not Abelian. The other ambiguity is that most activation functions commonly used are not equivariant with pooling operators.  However, since many of them are based on element-wise operations with monotonic functions, we can presume that the anomaly are not dominant in most cases. We find that this approach performs comparably to GMPool for small sized molecules where a single pooling depth suffice.

\section{Experiments}
\subsection{Experimental Setup}
\label{setup}
We arrange a total of five datasets to test our algorithms: two are open datasets collected from MoleculeNet \cite{Ramsundar-et-al-2019} and Binding DB \cite{10.2174/1386207013330670, 10.1093/bioinformatics/18.1.130, 10.1002/bip.10076, 10.1093/nar/gkl999, 10.1093/nar/gkv1072}, three are manually collected and arranged from different literatures including scientific articles and patents.
\begin{itemize}
    \item {\bf PLQY} includes experimentally measured values of photoluminescence quantum yield (PLQY) for fluorescence molecules. 
    \item {\bf $\lambda_{max}$ Solvents} contains measured $\lambda_{max}$, wavelength that shows maximum intensity for emission of a fluorescence molecule, under the solvent condition. 
    \item {\bf $\lambda_{max}$ Films} consists of $\lambda_{max}$ values measured after spin coating of fluorescence molecules on films doped with host materials. 
    \item {\bf pIC50} contains the negative log of the IC50 values for ATP receptor. IC50 implies minimum concentration of certain molecule needed for inhibiting half of activity of the target proteins. The IC50 values are optained from the BindingDB (https://www.bindingdb.org/bind/index.jsp). 
    \item {\bf Tox21} consists of results of 12 types of toxicity screening tests. We labeled a molecule `toxic' if the molecule failed in any of screening type. Data were originated from Tox21 challenge (2014). Since there are molecules without graph structure information in the dataset, we selected $7,831$ molecules that have the graph structure information.
\end{itemize}
For proper evaluation of pooling approaches, each graph in the data must have at least two or more effective groups.
However, Tox21 and pIC50 data contains molecules too small to contain multiple groups and thus we drop molecules with less than 20 nodes from the datasets. 
In addition, we drop molecules with more than 40 nodes from Tox21 and pIC50 datasets to accelerate the whole training process under dense matrix computations: the largest molecule in each respective dataset has 86 and 132 nodes, but the ratio of molecules with size over 40 in the dataset is only $3.4\%$ and $3.6\%$. Especially for pIC50 dataset, the proportion of molecules with less than 20 nodes are $0.4\%$. Lastly, the Tox21 task has been simplified to a single classification task by setting a positive label if any of the 12 tasks are positive in the original dataset. Details can be found in Table \ref{data_sum} and appendix section.
Every experiments are tested under five-fold settings with uniform sampling and 10\% of dedicated test set to secure the results, and single RTX 3090 is used for the experiments.

\subsection{Baselines}
For empirical evaluation, we compare the performance of GMPool and NGMPool against that of five other pooling approaches. We run all experiments via a pipeline with a fixed DMPNN backbone, while exchanging the pooling layers only. Here we provide brief descriptions of each baselines used: Top-k~\cite{gao2019} and SAGPool~\cite{lee2019} retain nodes with the highest scoring based on the projections of node features and self-attention scores, respectively. DiffPool~\cite{ying2018} uses an additional GNN to learn soft-assignment matrices that mix nodes into clusters. ASAPool~\cite{ranjan2020asap} clusters local subgraphs together through scoring and selection of clusters. MemPool~\cite{mempool} incorporates memory layers that jointly coarsen and transform input node representations. Note that we reimplemented the DMPNN backbone, Top-k pooling, and DiffPool. Implementations of other pooling baselines are borrowed from the \texttt{pytorch-geometric} library.

\begin{figure}[ht]
    \centering
    \begin{subfigure}[b]{0.3\textwidth}
        \centering
        \includegraphics[height=30mm, width=40mm, scale=0.13]{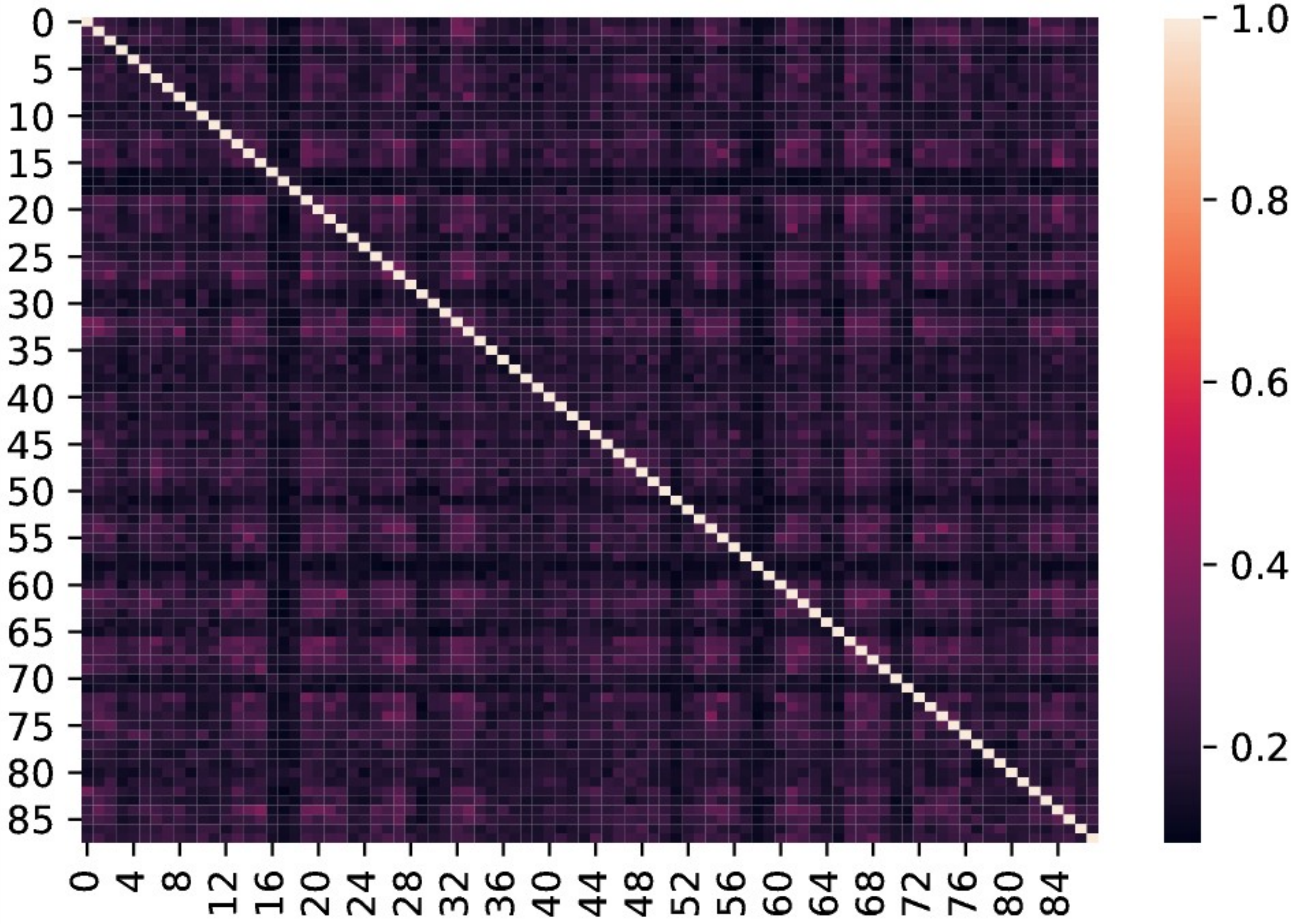}
        \caption{Initial State}
        \label{initial}
    \end{subfigure}
    \hspace{2mm}
    \begin{subfigure}[b]{0.3\textwidth}
        \centering
        \includegraphics[height=30mm, width=40mm, scale=0.13]{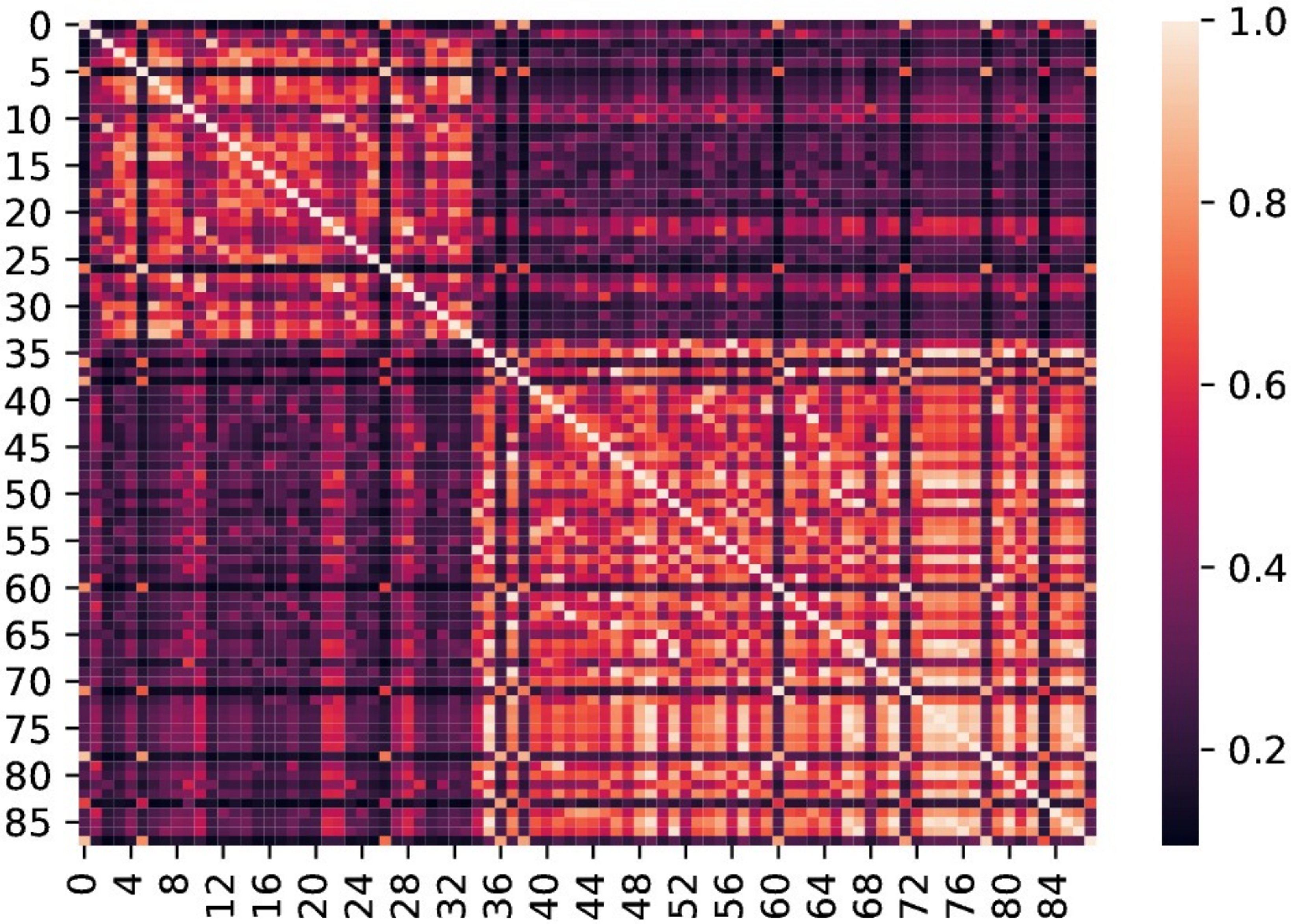}
        \caption{2-Groups}
        \label{simple}
    \end{subfigure}
    \hspace{2mm}
    \begin{subfigure}[b]{0.3\textwidth}
        \centering
        \includegraphics[height=30mm, width=40mm, scale=0.13]{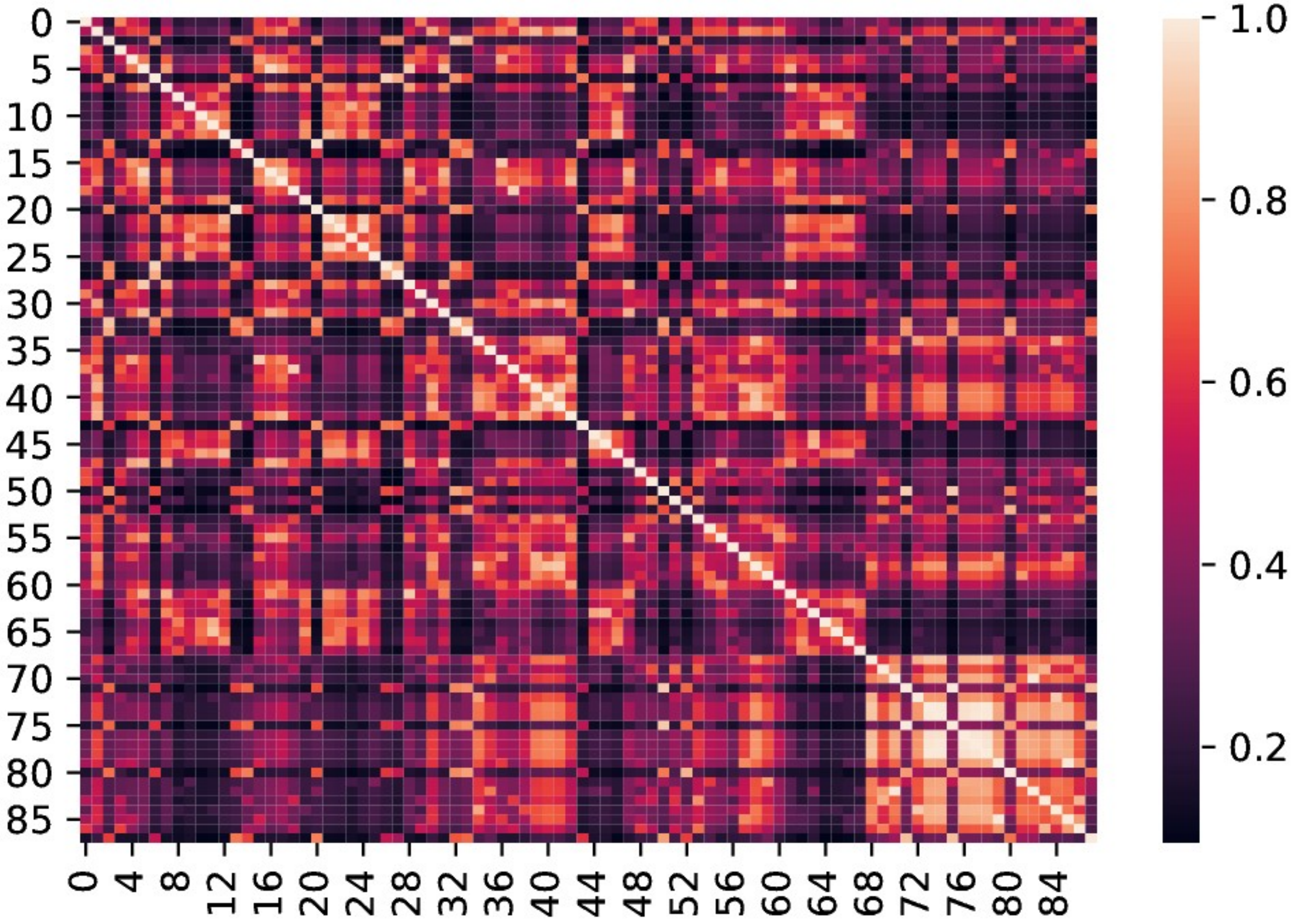}
        \caption{Multi-Groups}
        \label{comp}
    \end{subfigure}
    \caption{Examples of Grouping Matrices. Both $X$ and $Y$ axes are node indices. (a) is the initial state of grouping matrix, (b) and (c) show a grouping matrix for example molecules from PLQY and $\lambda_{max}$ datasets, respectively.}
    \label{grouping}
\end{figure}

\subsection{Hyperparameters}
\label{hyperparameters}
For the backbone of the model, DMPNN, we use the same hidden size of 200 across all three independent layers: the initial edge features with dimension $d_e$ and node features with dimension $d_n$ are passed through layers of dimension  $ d_e \times 200 $ and $ d_n \times 200 $, respectively with ReLU activation. The initial node and edge embeddings are determined by features generated in RDKit. The message passing module passes node embeddings through a linear layer with dimension $200 \times 200$, followed by ReLU activation and $0.15$ dropout layer. For graph representation we use a global average pooling scheme. GMPool and NGMPool construct the grouping matrix via a $200 \times 1$ linear layer and sigmoid activation without any parameters related to cluster numbers or thresholds. We use a batch size of $80$ and Adam optimizer for all model training.

For baseline pooling methods that require the cluster size as a hyperparameter, we perform grid search across candidates following previous work, and present best results. 
However, we fix the final pooling size to 10 as the average size of most common $40$ functional groups in bioactive molecules is $4.25$~\cite{ertl2020most}, indicating that molecules under concern (statistics shown in Table~\ref{data_sum}) can have up to $10$ clusters.
The specific hyperparameter setups used for pooling baselines can be found in appendix.

\subsection{Grouping Result}
The grouping matrix starts from randomized initial state and is optimized to gather effective functional groups in the molecules (Figures \ref{simple} and \ref{comp}).
Furthermore, since our algorithm fully enjoys the soft clustering concept, the result shows continuous weights for each group. This characteristic ensures the model can gather information from distant nodes if necessary. However, sometimes the grouping matrix shows unfamiliar forms, since the effective functional groups should vary due to the downstream task itself. For instance, for some simple tasks such as PLQY prediction, the grouping is rather intuitive as shown in Figure \ref{simple}, yet for complicated tasks like $\lambda_{\textrm{max}}$ prediction, the effective functional groups are also complicated as in Figure \ref{comp}.

\begin{table*}
  \caption{Test Results from Various Datasets}
  \label{test_result1}
  \centering
  \resizebox{\textwidth}{!}{\begin{tabular}{l|llllll|llll}
    \toprule
    \multirow{3}{*}{Models}    & \multicolumn{2}{c}{PLQY}    & \multicolumn{2}{c}{$\lambda_{max}$ Solvents}    &
    \multicolumn{2}{c}{$\lambda_{max}$ Films}  &
    \multicolumn{2}{|c}{pIC50}    & \multicolumn{2}{c}{Tox21}\\
     & \multicolumn{2}{c}{(RMSE $\downarrow$)}    & \multicolumn{2}{c}{(RMSE $\downarrow$)}    &
    \multicolumn{2}{c}{(RMSE $\downarrow$)}    &
    \multicolumn{2}{|c}{(RMSE $\downarrow$)}    & \multicolumn{2}{c}{(ROC-AUC $\uparrow$)}\\
    \cmidrule(r){2-3}
    \cmidrule(r){4-5}
    \cmidrule(r){6-7}
    \cmidrule(r){8-9}
    \cmidrule(r){10-11}
         & Mean    & Std    & Mean    & Std    & Mean    & Std & Mean    & Std    & Mean    & Std \\
    \midrule
    GCN & 0.2619   & 0.0519    & 64.09    & 2.281    & 58.33    & 2.419 & 0.7390    & 0.0265    & 0.6902    & 0.0305 \\
    DMPNN     & 0.2414  & 0.0093    & 46.22    & 0.523    & 35.10    & 1.205 & 0.7839    & 0.0076   & 0.7191    & 0.0138 \\
    Top-k & 0.2204        & 0.0143    & 47.06    & 2.086     & 50.60    & 0.765 & 0.7860    & 0.0107    & 0.7593    & 0.0160\\
    SAGPool & 0.2123        & 0.0166    & 45.44    & 3.812     & 37.98    & 1.968 & 0.7956        & 0.0072    & 0.7155    & 0.0342  \\
    DiffPool & 0.2386        & 0.0181    & 56.54    & 0.189     & 52.78    & 0.335 & 0.8091        & 0.0126    & 0.6619    & 0.0155  \\
    ASAPool & 0.2204        & 0.0211    & 53.19    & 3.548     & 47.39    & 2.434 & 0.8310        & 0.0173    & 0.6864    & 0.0466  \\
    MemPool & 0.2188        & 0.0081    & 44.48    & 2.010     & 38.54    & 3.249 & 0.7910        & 0.0073    & 0.6926    & 0.0257  \\
    \midrule
    GMPool     & {\bf 0.2091}       & 0.0081    & 44.48    & 2.010    & {\bf 34.34}    & 3.027 & {\bf 0.6951}    & 0.0270  & 0.7451    & 0.0197\\
    NGMPool     & 0.2169       & 0.0340    & {\bf 38.61}    & 2.573    & 39.82    & 0.702 & 0.7253    & 0.0198  & {\bf 0.7617}    & 0.0201\\
    \bottomrule
  \end{tabular}
  }
\end{table*}

\subsection{Main Result}

We tested various combinations of models and dataset to check the validity of our algorithm. We selected GCN, DMPNN, Top-k, SAGPool, DiffPool, ASAPool and MemPool algorithm to set a benchmark score to compare with. As it is shown in the table \ref{test_result1}, majority of the cases, our models outperform conventional methods. However, for some tasks (i.e. $\lambda_{\textrm{max}}$ datasets), our model is gaining only a small margin of the performance. This is caused by the underlying mechanism of the chemical effect. Since some tasks are strongly related to the effective groups of the molecule, yet others are not. In those cases, sub-structures are not intuitive and might appear in very complicated forms, as shown in Figure \ref{comp}. If the grouping becomes complicated, the rank of the pooling matrix should be larger to cover all degrees of freedom for the data. However, conventional models, which shared predefined numbers as universal grouping numbers, force to collect groups and reduce it to the low-rank form, which might not have enough degree of freedom. This will cause information loss or blend which compromises the prediction result. Therefore, one can check that in $\lambda_{\textrm{max}}$ prediction test, conventional pooling algorithms show inferior result than simple massage passing scheme. Yet our model is not designed to reduce the physical rank of the matrix during the pooling process, and there is always enough degree of freedom to carry the information throughout learning. Hence, even for the $\lambda_{\textrm{max}}$ case, our model outperforms the others. Furthermore, for other tasks, it is clear that our model improves performance by $5 \sim 10\%p$.
\begin{figure}[ht]
    \centering
    \begin{subfigure}[c]{0.42\textwidth}
    \centering
    \hspace{-10mm}
        \includegraphics[height=43mm, width=55mm, scale=0.25]{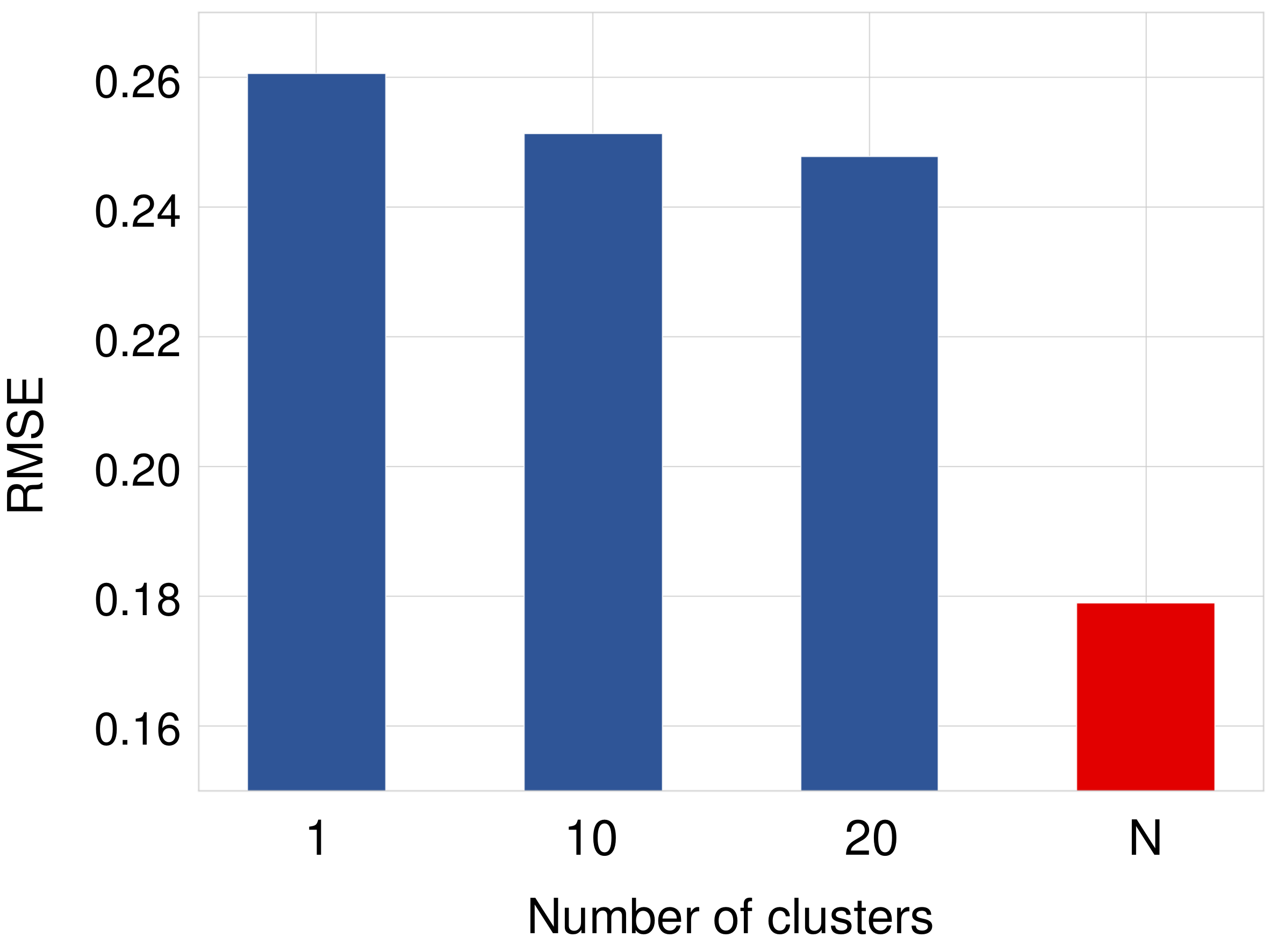}
        \caption{}
        \label{ab1}
    \end{subfigure}
    \hspace{10mm}
    \begin{subfigure}[c]{0.42\textwidth}
        \includegraphics[height=43mm, width=55mm, scale=0.25]{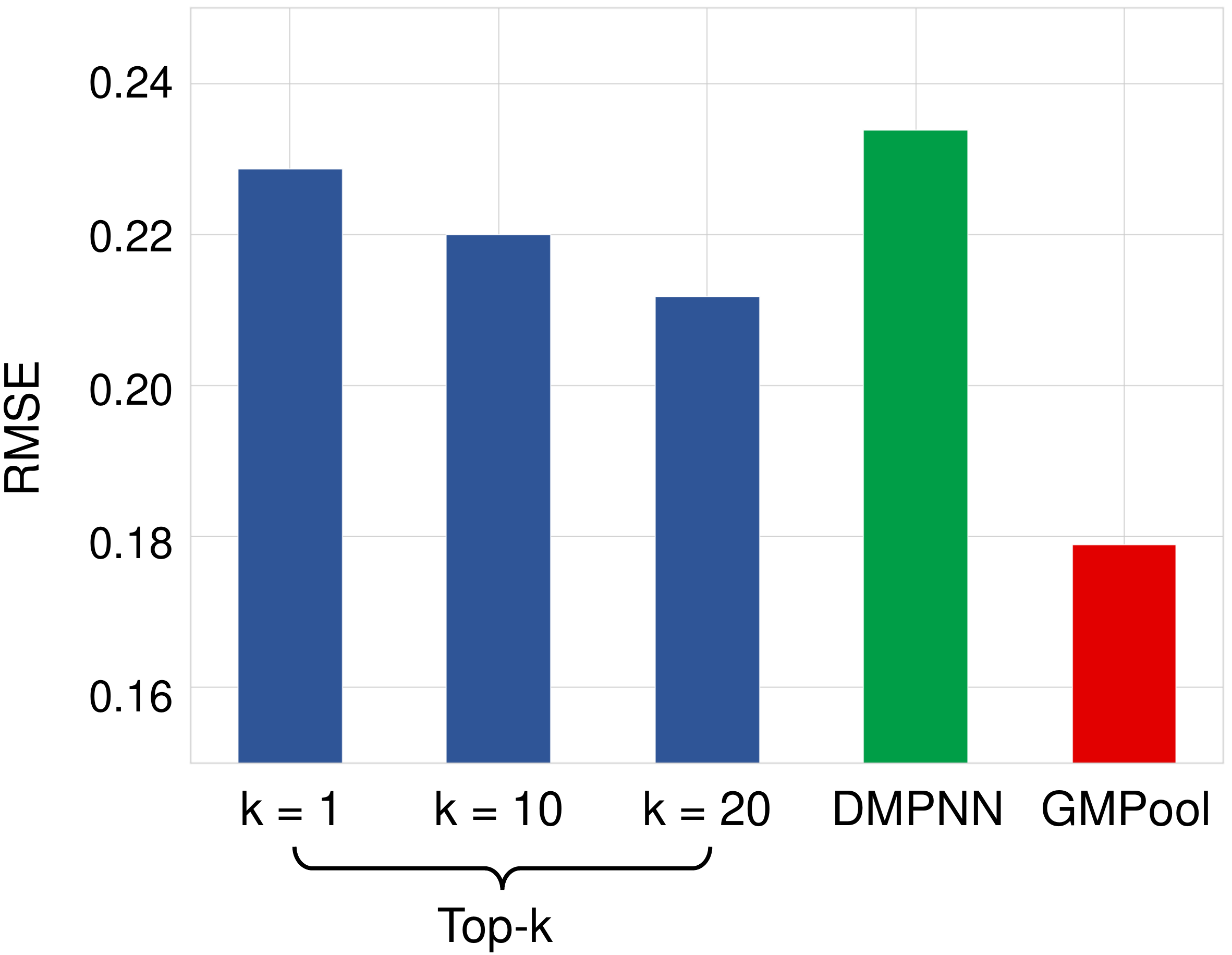}
        \caption{}
        \label{ab2}
    \end{subfigure}
\caption{RMSE for PLQY prediction (a) over varying number of pooling layers in \methodA{} and (b) across different methods. For figure (a), \methodA{} does not require to set number of clusters, however to explore the effect, the number of clusters is set by force in the first place.}
\label{ablation}
\end{figure}

\subsection{Ablation Study}


One crucial hyperparameter to be explored in pooling models is the number of clusters. Even though our model does not require to fix the number of clusters in the first place, one can set the parameter by force. One can easily see in the Figure \ref{ab1} that the number of clusters can be set to the number of nodes without compromising performance of the model. Further, Figure \ref{ab2} shows that our model outperforms Top k algorithms with various cluster numbers and original DMPNN as well. This is one of the powerful features of our model, since the model automatically splits groups and determines the appropriate number of sub-structures for each individual graph. One can also force the number of clusters and share through all graphs in an equal manner; however, it is not effective for the following reasons. In real world data, one can not esteem the exact number of clusters for individual graphs. This might be problematic if one sets the number of clusters less than it requires, the models' performance will be compromised due to the information loss. Another problem is caused by the mathematical structure of the decomposition scheme. Using SVD method will cause ambiguity since collecting only top k eigen values from the decomposed matrix might not reconstruct the original grouping matrix due to lack of information. It is even worse in the initial stage of the learning as the weight is almost in the random state and the top k eigen values are not precisely representing the appropriate clusters. Thus, as it is depicted in the above figure, it is best to leave the cluster number to be determined automatically by the model itself.

\section{Conclusion}
\label{conclusion}
We have introduced a novel pooling architecture with adaptive number of clusters based on a second order pooling operator, namely the grouping matrix. The grouping matrix is based on clustering similarities between every possible pairs of nodes, ensuring permutation invariance. We have shown that our model is valid for chemical property prediction and outperforms conventional methods in real-world datasets.

While our model is useful and effective, there is still room for improvement. First of all, despite leveraging a method to decompose the grouping matrix with stable gradient computations, there exist corner cases with a small eigengap at which the model fails to converge. This event seldom happens (about $0.00018\%$ in our experiments), but can be non-negligible when one needs to learn with a large number of data points. Hence, one future direction would be to impose proper constraints on the loss to avoid such gradient blowup in the grouping matrix. 

Another future direction would be to enhance scalability of our methods to improve applicability to large-scale graphs. Since the grouping matrix decomposition step via SVD is the main computational bottleneck of \methodA{}, incorporating faster decomposition modules such as randomized approximation~\cite{halko2011finding, DBLP:journals/corr/abs-1710-02812} methods can lead to faster inference. However, this is likely to incur loss in predictive performance, and as the focus of this work lies in allowing variation in the number of clusters in small molecular graphs where scalability is not an issue, we defer improving the scalability to future work.

Lastly, generalizing the second order grouping matrix towards higher-order grouping tensors can allow further expressive power. We have introduced a pairwise structure; yet it is not obliged to be fixed into the pairwise form. If we consider higher-order form of node combinations, i.e. k-form where $k < N$ and $N$ is total node number, the grouping matrix can be generalized into the higher rank tensor. Based on the tensor-form, the transformation rule can be written as

\begin{equation}
    \tilde{M}_{\mu_1 \cdots \mu_k} = S_{\mu_1}^{\phantom{\mu_1}\nu_1} \cdots S_{\mu_k}^{\phantom{\mu_k}\nu_k}M_{\nu_1 \cdots \nu_k}
\end{equation}

Note that to satisfy the above transformation rule, the following conditions are required. One is that selecting nodes combination should be as same as selecting nodes {\it set} and size of the {\it set} should fixed into the number of nodes in a group. The other is that the classification result of the node set should be retained the same for any subset in the node set. This concept may have a connection to hypergraph configurations. However if we raise the nodes numbers above $2$, required computation power increases by a huge amount, since the combination number grows exponentially until the number of nodes hits $N/2$. Therefore, practically it is a difficult task to test the higher rank version of our algorithm, yet it could be useful for learning datasets with higher order connections.

\bibliographystyle{abbrv}
\bibliography{references}

\begin{thebibliography}{10}

\bibitem{https://doi.org/10.48550/arxiv.1609.02907}
Thomas~N. Kipf and Max Welling.
\newblock Semi-supervised classification with graph convolutional networks.
\newblock 2016.

\bibitem{https://doi.org/10.48550/arxiv.1706.02216}
William~L. Hamilton, Rex Ying, and Jure Leskovec.
\newblock Inductive representation learning on large graphs.
\newblock 2017.

\bibitem{https://doi.org/10.48550/arxiv.1509.09292}
David Duvenaud, Dougal Maclaurin, Jorge Aguilera-Iparraguirre, Rafael
  Gómez-Bombarelli, Timothy Hirzel, Alán Aspuru-Guzik, and Ryan~P. Adams.
\newblock Convolutional networks on graphs for learning molecular fingerprints.
\newblock 2015.

\bibitem{https://doi.org/10.48550/arxiv.1606.09375}
Michaël Defferrard, Xavier Bresson, and Pierre Vandergheynst.
\newblock Convolutional neural networks on graphs with fast localized spectral
  filtering.
\newblock 2016.

\bibitem{https://doi.org/10.48550/arxiv.1312.6203}
Joan Bruna, Wojciech Zaremba, Arthur Szlam, and Yann LeCun.
\newblock Spectral networks and locally connected networks on graphs.
\newblock 2013.

\bibitem{doi:10.1021/acs.jcim.6b00601}
Connor~W. Coley, Regina Barzilay, William~H. Green, Tommi~S. Jaakkola, and
  Klavs~F. Jensen.
\newblock Convolutional embedding of attributed molecular graphs for physical
  property prediction.
\newblock {\em Journal of Chemical Information and Modeling}, 57(8):1757--1772,
  2017.
\newblock PMID: 28696688.

\bibitem{4700287}
Franco Scarselli, Marco Gori, Ah~Chung Tsoi, Markus Hagenbuchner, and Gabriele
  Monfardini.
\newblock The graph neural network model.
\newblock {\em IEEE Transactions on Neural Networks}, 20(1):61--80, 2009.

\bibitem{https://doi.org/10.48550/arxiv.1812.01070}
Wengong Jin, Kevin Yang, Regina Barzilay, and Tommi Jaakkola.
\newblock Learning multimodal graph-to-graph translation for molecular
  optimization, 2018.

\bibitem{ying2018}
Rex Ying, Jiaxuan You, Christopher Morris, Xiang Ren, William~L. Hamilton, and
  Jure Leskovec.
\newblock Hierarchical graph representation learning with differentiable
  pooling.
\newblock {\em CoRR}, abs/1806.08804, 2018.

\bibitem{gao2019}
Hongyang Gao and Shuiwang Ji.
\newblock Graph u-nets.
\newblock {\em CoRR}, abs/1905.05178, 2019.

\bibitem{lee2019}
Junhyun Lee, Inyeop Lee, and Jaewoo Kang.
\newblock Self-attention graph pooling.
\newblock {\em CoRR}, abs/1904.08082, 2019.

\bibitem{bianchi2019}
Filippo~Maria Bianchi, Daniele Grattarola, and Cesare Alippi.
\newblock Mincut pooling in graph neural networks.
\newblock {\em CoRR}, abs/1907.00481, 2019.

\bibitem{ranjan2020asap}
Ekagra Ranjan, Soumya Sanyal, and Partha Talukdar.
\newblock Asap: Adaptive structure aware pooling for learning hierarchical
  graph representations.
\newblock In {\em Proceedings of the AAAI Conference on Artificial
  Intelligence}, volume~34, pages 5470--5477, 2020.

\bibitem{https://doi.org/10.48550/arxiv.1810.00826}
Keyulu Xu, Weihua Hu, Jure Leskovec, and Stefanie Jegelka.
\newblock How powerful are graph neural networks?
\newblock 2018.

\bibitem{https://doi.org/10.48550/arxiv.1710.10903}
Petar Veličković, Guillem Cucurull, Arantxa Casanova, Adriana Romero, Pietro
  Liò, and Yoshua Bengio.
\newblock Graph attention networks.
\newblock 2017.

\bibitem{https://doi.org/10.48550/arxiv.1704.01212}
Justin Gilmer, Samuel~S. Schoenholz, Patrick~F. Riley, Oriol Vinyals, and
  George~E. Dahl.
\newblock Neural message passing for quantum chemistry.
\newblock 2017.

\bibitem{doi:10.1021/acs.jcim.9b00237}
Kevin Yang, Kyle Swanson, Wengong Jin, Connor Coley, Philipp Eiden, Hua Gao,
  Angel Guzman-Perez, Timothy Hopper, Brian Kelley, Miriam Mathea, Andrew
  Palmer, Volker Settels, Tommi Jaakkola, Klavs Jensen, and Regina Barzilay.
\newblock Analyzing learned molecular representations for property prediction.
\newblock {\em Journal of Chemical Information and Modeling}, 59(8):3370--3388,
  2019.
\newblock PMID: 31361484.

\bibitem{NIPS2011_6c1da886}
Adam Coates and Andrew Ng.
\newblock Selecting receptive fields in deep networks.
\newblock In J.~Shawe-Taylor, R.~Zemel, P.~Bartlett, F.~Pereira, and K.Q.
  Weinberger, editors, {\em Advances in Neural Information Processing Systems},
  volume~24. Curran Associates, Inc., 2011.

\bibitem{10.1016/j.patcog.2006.04.007}
Dan Kushnir, Meirav Galun, and Achi Brandt.
\newblock Fast multiscale clustering and manifold identification.
\newblock {\em Pattern Recogn.}, 39(10):1876–1891, oct 2006.

\bibitem{https://doi.org/10.48550/arxiv.0711.0189}
Ulrike von Luxburg.
\newblock A tutorial on spectral clustering.
\newblock 2007.

\bibitem{4302760}
Inderjit~S. Dhillon, Yuqiang Guan, and Brian Kulis.
\newblock Weighted graph cuts without eigenvectors a multilevel approach.
\newblock {\em IEEE Transactions on Pattern Analysis and Machine Intelligence},
  29(11):1944--1957, 2007.

\bibitem{https://doi.org/10.1111/cbdd.12952}
Vishnupriya Kanakaveti, Ramasamy Sakthivel, S.~K. Rayala, and M.~Michael
  Gromiha.
\newblock Importance of functional groups in predicting the activity of small
  molecule inhibitors for bcl-2 and bcl-xl.
\newblock {\em Chemical Biology \& Drug Design}, 90(2):308--316, 2017.

\bibitem{doi:10.1021/acs.jmedchem.0c00754}
Peter Ertl, Eva Altmann, and Jeffrey~M. McKenna.
\newblock The most common functional groups in bioactive molecules and how
  their popularity has evolved over time.
\newblock {\em Journal of Medicinal Chemistry}, 63(15):8408--8418, 2020.
\newblock PMID: 32663408.

\bibitem{GUVENCH20161928}
Olgun Guvench.
\newblock Computational functional group mapping for drug discovery.
\newblock {\em Drug Discovery Today}, 21(12):1928--1931, 2016.

\bibitem{DBLP:journals/corr/IonescuVS15}
Catalin Ionescu, Orestis Vantzos, and Cristian Sminchisescu.
\newblock Training deep networks with structured layers by matrix
  backpropagation.
\newblock {\em CoRR}, abs/1509.07838, 2015.

\bibitem{DBLP:journals/corr/abs-1906-09023}
Wei Wang, Zheng Dang, Yinlin Hu, Pascal Fua, and Mathieu Salzmann.
\newblock Backpropagation-friendly eigendecomposition.
\newblock {\em CoRR}, abs/1906.09023, 2019.

\bibitem{9400752}
Wei Wang, Zheng Dang, Yinlin Hu, Pascal Fua, and Mathieu Salzmann.
\newblock Robust differentiable svd.
\newblock {\em IEEE Transactions on Pattern Analysis and Machine Intelligence},
  pages 1--1, 2021.

\bibitem{DBLP:journals/corr/abs-2105-02498}
Yue Song, Nicu Sebe, and Wei Wang.
\newblock Why approximate matrix square root outperforms accurate {SVD} in
  global covariance pooling?
\newblock {\em CoRR}, abs/2105.02498, 2021.

\bibitem{DBLP:journals/corr/abs-2201-08663}
Yue Song, Nicu Sebe, and Wei Wang.
\newblock Fast differentiable matrix square root.
\newblock {\em CoRR}, abs/2201.08663, 2022.

\bibitem{Ramsundar-et-al-2019}
Bharath Ramsundar, Peter Eastman, Patrick Walters, Vijay Pande, Karl Leswing,
  and Zhenqin Wu.
\newblock {\em Deep Learning for the Life Sciences}.
\newblock O'Reilly Media, 2019.
\newblock
  \url{https://www.amazon.com/Deep-Learning-Life-Sciences-Microscopy/dp/1492039837}.

\bibitem{10.2174/1386207013330670}
Chen X, Liu M, and Gilson MK.
\newblock {indingDB: a web-accessible molecular recognition database.}
\newblock {\em Combinatorial Chemistry \& High Throughput Screening},
  4(8):719--725, 12 2001.

\bibitem{10.1093/bioinformatics/18.1.130}
Chen X, Lin Y, Liu M, and Gilson MK.
\newblock {The Binding Database: data management and interface design.}
\newblock {\em Bioinformatics}, 18(1):130--139, 1 2002.

\bibitem{10.1002/bip.10076}
X.~Chen, Y.~Lin, and M.K. Gilson.
\newblock {The Binding Database: Overview and User's Guide}.
\newblock {\em Biopolymers}, 61:127--141, 2002.

\bibitem{10.1093/nar/gkl999}
Liu T, Lin Y, Wen X, Jorissen RN, and Gilson MK.
\newblock {BindingDB: a web-accessible database of experimentally determined
  protein-ligand binding affinities.}
\newblock {\em Nucleic Acids Research}, 35:D198--D202, 1 2007.

\bibitem{10.1093/nar/gkv1072}
Michael~K. Gilson, Tiqing Liu, Michael Baitaluk, George Nicola, Linda Hwang,
  and Jenny Chong.
\newblock {BindingDB in 2015: A public database for medicinal chemistry,
  computational chemistry and systems pharmacology}.
\newblock {\em Nucleic Acids Research}, 44(D1):D1045--D1053, 10 2015.

\bibitem{mempool}
Amir~Hosein Khasahmadi, Kaveh Hassani, Parsa Moradi, Leo Lee, and Quaid Morris.
\newblock Memory-based graph networks, 2020.

\bibitem{ertl2020most}
Peter Ertl, Eva Altmann, and Jeffrey~M McKenna.
\newblock The most common functional groups in bioactive molecules and how
  their popularity has evolved over time.
\newblock {\em Journal of medicinal chemistry}, 63(15):8408--8418, 2020.

\bibitem{halko2011finding}
Nathan Halko, Per-Gunnar Martinsson, and Joel~A Tropp.
\newblock Finding structure with randomness: Probabilistic algorithms for
  constructing approximate matrix decompositions.
\newblock {\em SIAM review}, 53(2):217--288, 2011.

\bibitem{DBLP:journals/corr/abs-1710-02812}
Vinita Vasudevan and M.~Ramakrishna.
\newblock A hierarchical singular value decomposition algorithm for low rank
  matrices.
\newblock {\em CoRR}, abs/1710.02812, 2017.

\bibitem{Sedov_2018}
Denis Sedov and Zhirong Yang.
\newblock Word embedding based on low-rank doubly stochastic matrix
  decomposition.
\newblock In {\em Neural Information Processing}, pages 90--100. Springer
  International Publishing, 2018.

\end{thebibliography}



\newpage

\appendix
\section{Appendix}
\subsection{Notation for Appendix}
Our notation is based on index notation and Einstein summation conventions. Notation of functions and matrices in our algorithm is as follows.

\begin{gather*}
    X_{\mu \nu \cdots}^{h_i} : \textrm{Input Data} \\ 
    W_{h_i}^{\phantom{h_i}h_j} : \textrm{Weight of layer}\\
    f(\cdot) : \textrm{Activation Function} \\ 
    S_{\phantom{\mu}\nu}^{\mu} : \textrm{Pooling Transformation Operator}\\
    M_{\mu\nu} : \textrm{Grouping Matrix} \\
    D_{\mu\nu} : \textrm{Degree Matrix}\\
    F_{\textrm{(node)}\mu}^{\phantom{node \mu}h_i} : \textrm{Node Features} \\ 
    F_{\textrm{(edge)}\mu\nu}^{\phantom{edge \mu\nu}h_i} : \textrm{Edge Features}\\
    A_{\mu\nu} : \textrm{Adjacency Matrix} \\
    \Gamma_{\mu\nu}^{\phantom{\mu\nu}\rho} : \textrm{Generalized Adjacency Tensor}
\end{gather*}
There are types of indices according to the representation space.
\begin{gather*}
    \mu, \nu : \textrm{Tensor Component Indices}\\
    h_i : \textrm{Indices for Hidden Dimension}
\end{gather*}

However, we assume all representation spaces are in Euclidean. Therefore, all indices are raised and lowered by Euclidean metric $I_{ij}$.


\subsection{Detailed Description on Grouping Matrix}
The grouping matrix is based on a binary classification scheme of pairwise node combinations. Therefore, we start by acquiring a node representation by GNN model. Here we use DMPNN as an encoder since its performance on molecular property prediction is qualified. To form the grouping matrix, the rank 2 tensor of the pairwise nodes combination should be computed first. This tensor should be symmetric since the order of the node is irrelevant, hence any commutative functions can be used in the procedure. Here we focus on the Euclidean distance between nodes pair to simplify the classification.
\begin{eqnarray}
\label{input}
    X_{(\mu\nu)}^{h_i} = |\Gamma_{\mu\nu}^{\phantom{\mu\nu}\gamma}F_{\textrm{(node)}\gamma^{\phantom{\mu}h_i}} - (\Gamma_{\mu\nu}^{\phantom{\mu\nu}\gamma}F_{\textrm{(node)}_\gamma^{\phantom{\gamma}h_i}})^T|
\end{eqnarray}

Where $\Gamma_{\mu\nu}^{\phantom{\mu\nu}\rho}$ is defined as follows.

\begin{equation}
    \Gamma_{\mu\nu}^{\phantom{\mu\nu}\rho} = 
    \Bigg{\{}
    \begin{array}{l}
         1 \qquad \textrm{if} \quad \nu = \rho \\
         \\
         0 \qquad \textrm{otherwise}
    \end{array}
\end{equation}

The binary classifier is set to be simple enough to avoid overfitting issues. In our case, we used a single layer with softmax activation to extract probability for each pair of nodes.

\begin{equation}
\label{gm}
    M_{(\mu\nu)} = f(W_{h_i}X_{\mu\nu}^{h_i})
\end{equation}

Since $X_{\mu \nu}$ is symmetric under indices, our grouping matrix is also manifestly symmetric. Every step of computation is based on node level, and if activation function $f$ is an element-wise operation, it is clear to see that eq. \ref{input}, eq. \ref{gm} are permutation equivariant with permutation operator $P_\mu^{\phantom{\mu}\nu}$.

\begin{eqnarray}
   \tilde{X}_{(\rho\lambda)}^{h_i}& = & P_\rho^{\phantom{\rho}\mu}X_{(\mu\nu)}^{h_i}(P^T)^\nu_{\phantom{\nu}\lambda} \nonumber\\
   &=& P_\rho^{\phantom{\rho}\mu}|\Gamma_{\mu\nu}^{\phantom{\mu\nu}\gamma}F_{\textrm{(node)}\gamma^{\phantom{\mu}h_i}} \nonumber\\ && - (\Gamma_{\mu\nu}^{\phantom{\mu\nu}\gamma}F_{\textrm{(node)}\gamma^{\phantom{\gamma}h_i}})^T|(P^T)^\nu_{\phantom{\nu}\lambda} \\ 
   \tilde{M}_{(\rho\lambda)} &=& 
   P_\rho^{\phantom{\rho}\mu}M_{(\mu\nu)}(P^T)^\nu_{\phantom{\nu}\lambda} \nonumber\\
   & =& P_\rho^{\phantom{\rho}\mu}f(W_{h_i}X_{\mu\nu}^{h_i})(P^T)^\nu_{\phantom{\nu}\lambda}
\end{eqnarray}
For the ideal grouping case that no nodes pairs are assigned in multiple groups simultaneously, the grouping matrix contains only $0, 1$ for its elements. With proper permutation, it is always possible to gather nodes pairs in the same groups in close range and make the grouping matrix to be a block diagonal. Block diagonal matrix is simple to interpret. The number of blocks corresponds to the number of groups after pooling and nodes pairs assigned to the same groups should be gathered during pooling operation. For instance, if there are three different groups and each group size are $k_1, k_2, k_3$,
\begin{equation}
  \setlength{\arraycolsep}{0pt}
  M_{\mu\nu} = \begin{bmatrix}
    \fbox{$1_{k_1 \times k_1}$} & 0 & 0  \\
    0 & \fbox{$1_{k_2 \times k_2}$} & 0 \\
    0 & 0 & \fbox{$1_{k_3 \times k_3}$} \\
  \end{bmatrix}
  \label{groupingmatrix}
\end{equation}
One can easily see that pooling operator is in following form.
\begin{equation}
  \setlength{\arraycolsep}{0pt}
  S^{\mu}_{\phantom{\mu}i} = \begin{bmatrix}
    \fbox{$1_{k_1 \times 1}$} & 0 & 0 \quad & \cdots & \quad 0\\
    0 & \fbox{$1_{k_2 \times 1}$} & 0 \quad & \cdots & \quad 0\\
    0 & 0 & \fbox{$1_{k_3 \times 1}$} \quad & \cdots & \quad 0\\
  \end{bmatrix}
\end{equation}
In general cases, every single element should be generalized to a continuous number in the range of $[0, 1]$ for the grouping matrix in eq. \ref{groupingmatrix} which indicates the soft clustering case.

\begin{table*}
  \caption{Hyperparameters on Models}
  \label{model_oriented}
  \centering
  \begin{tabular}{l|lllll}
    \toprule
    Models     & Layer \#    & Hops  & Pooling \#   & Pooling Depth & Heads\\
    \midrule
    GCN & 3 & 4    & -    & -  & -\\
    DMPNN     & 3   & 4    & -   & -  & -\\
    Top-k & 1   & [4, 2]    & [10]    & 1  & -\\
    SAGPool & 3   &[4, 1, 1]     & [0.5, 10]    & 2  & -\\
    DiffPool & 2   &[2, 1, 1, 1]     & [0.7, 0.5, 10]    & 3  & -\\
    ASAPool & 3     &[4, 1, 1]   & [0.5, 10]    & 2  & -\\
    MemPool & -     &[4, 2]   & [10]    & 1  & 5\\
    \midrule
    GMPool    & 1   &[4, 2] & -  & 1  & -\\
    NGMPool    & 1  &[4, 2]  & -  & 1  & -\\
    \bottomrule
  \end{tabular}
\end{table*}

\subsection{Detailed Proof of Eq. 16 and 17}
\label{detail}
The key idea is to retain pooling depth to one and using transformed aggregation vectors to pooled space as aggregation basis. Base model we used is DMPNN. It is convenient to start with DMPNN algorithm in algebraic form.

\begin{eqnarray}
    \label{mpn_bond} ({F'_{\textrm{(edge)}}})_{\mu\nu}^{\phantom{\mu\nu}h_i} &=& f(W_{h_i}^{\phantom{h_i}h_j} (\Gamma_{\mu\nu}^{\phantom{\mu\nu}\rho} ({F_{\textrm{(edge)}}}_{\rho\lambda}1^\lambda) \nonumber\\  && 
    - ({F_{\textrm{(edge)}}}^T)_{\mu\nu})_{h_j})\\
    \label{mpn_atom} ({F'_{\textrm{(node)}}})_{\mu}^{\phantom{\mu}h_i} &=& f({F'_{\textrm{(edge)}}}_{\mu\nu}^{h_i} 1^\nu )
\end{eqnarray}
Suppose the pooling operator $S_\mu^{\phantom{\mu}\nu}$ is given, one can easily derive pooling process for above node and edge features.
\begin{equation} \label{transrel}
\begin{array}{ccl}
    ({\tilde{F}'_{\textrm{(edge)}}})_{\mu\nu}^{\phantom{\mu\nu}h_i} &=& S_{\mu}^{\phantom{\mu}\rho} {F'_{\textrm{(edge)}}}_{\rho\lambda}^{\phantom{\rho\lambda}h_i} (S^T)^{\lambda}_{\phantom{\lambda}\nu}\\
    ({\tilde{F}'_{\textrm{(node)}}})_{\mu}^{\phantom{\mu}h_i} &=& S_{\mu}^{\phantom{\mu}\rho} {F'_{\textrm{(node)}}}_{\rho}^{\phantom{\rho}h_i}\\
    (\tilde{\Gamma}')_{\mu\nu}^{\phantom{\mu\nu}\rho} &=& S_{\mu}^{\phantom{\mu}\gamma} \Gamma_{\gamma \lambda}^{\phantom{\gamma\lambda}\delta}(S^T)^{\lambda}_{\phantom{\lambda}\nu}(S^T)_{\delta}^{\phantom{\delta}\rho}\\
    \tilde{1}'_{\mu} &=& S_{\mu}^{\phantom{\mu}\nu} 1_{\nu}
\end{array}
\end{equation}
Also, the message passing process after pooling by utilizing eq. \ref{mpn_bond}, \ref{mpn_atom} and \ref{transrel} can be computed as follows.
\begin{eqnarray}
    G_{\sigma\delta h_j} &=& (\Gamma_{\sigma\delta}^{\phantom{\sigma\delta}\rho} M_\rho^{\phantom{\rho}\alpha}({F'_{\textrm{(edge)}}}_{\alpha\beta}M^\beta_{\phantom{\beta}\gamma}1^\gamma) \\ \nonumber &&- ({F_{\textrm{(edge)}}}^T)_{\sigma\delta})_{h_j}\\
    ({\tilde{F}''_{\textrm{(edge)}}})_{\mu\nu}^{\phantom{\mu\nu}h_i} &=& f(W_{h_i}^{\phantom{h_i}h_j}S_{\mu}^{\phantom{\mu}\sigma} G_{\sigma\delta h_j} (S^T)^{\delta}_{\phantom{\sigma}\nu})\\
    ({\tilde{F}''_{\textrm{(node)}}})_{\mu}^{\phantom{\mu}h_i} &=& f(S_\mu^{\phantom{\mu}\nu}{F'_{\textrm{(edge)}}}_{\nu\rho}^{h_i} M^\rho_{\phantom{\rho}\lambda}1^\lambda )
\end{eqnarray}

If we assume the anomaly caused by commuting activation function with pooling operation is not dominant, then the above relation can be approximated in the following form.

\begin{eqnarray}
    ({\tilde{F}''_{\textrm{(edge)}}})_{\mu\nu}^{\phantom{\mu\nu}h_i} &\sim& S_{\mu}^{\phantom{\mu}\sigma}f(W_{h_i}^{\phantom{h_i}h_j} G_{\sigma \delta h_j})(S^T)^{\delta}_{\phantom{\delta}\nu}\\
    ({\tilde{F}''_{\textrm{(node)}}})_{\mu}^{\phantom{\mu}h_i} &\sim& S_\mu^{\phantom{\mu}\nu}f({F'_{\textrm{(edge)}}}_{\nu\rho}^{h_i}) M^\rho_{\phantom{\rho}\lambda}1^\lambda
\end{eqnarray}

Finally, after aggregation to acquire graph embedding, the transformation operator can be replaced by a grouping matrix.

\begin{eqnarray*}
    \tilde{1}^\mu({\tilde{F}''_{\textrm{(edge)}}})_{\mu\nu}^{\phantom{\mu\nu}h_i}\tilde{1}^\nu &=& 1^\mu M_\mu^{\phantom{\mu}\gamma}f(W_{h_i}^{\phantom{h_i}h_j}(G_{\sigma \delta h_j})M^{\sigma}_{\phantom{\sigma}\nu} 1^\nu\\
    \tilde{1}^\mu({\tilde{F}''_{\textrm{(node)}}})_{\mu}^{\phantom{\mu}h_i} &=& 1^\mu M_\mu^{\phantom{\mu}\nu}f({F'_{\textrm{(edge)}}}_{\nu\rho}^{h_i}) M^\rho_{\phantom{\rho}\lambda}1^\lambda
\end{eqnarray*}


\subsection{Numerical Iterative Decomposition Scheme}
Another interesting decomposition method is the numerical iterative scheme. Standard SVD is theoretically robust and simple. However, there is some ambiguity if one tries to pick top-k eigenvalues and reduce the transformation operator into low-rank form, since the low-rank pooling operator might not be able to reconstruct the original grouping matrix. Hence, if one desires to acquire low-rank pooling operators directly, SVD method is not the one to choose. There are some useful works about decomposing matrix by numerical iterative scheme. \cite{Sedov_2018} One that we introduce is rather simple. Set a low-rank weight and demand to recover the original given matrix by taking the square of the weight.
\begin{eqnarray}
    \label{obj_nd}\textrm{min} \, \cal{L}(W) = D(M||\hat{M})\\
    \label{con1}\textrm{where} \quad \hat{M} = \frac{(W^T) W}{\sum_i W_{ij}}\\
    \label{con2}\textrm{and} \quad {\sum_i W_{ij}} = 1
\end{eqnarray}
Eq. \ref{obj_nd} denotes loss function to optimize. In eq. \ref{con1}, the matrix $M$ is the given grouping matrix and $\hat{M}$ is the reconstructed matrix from low-rank approximated weight $W$. Finally eq. \ref{con2} insist to have sum of elements over reduced rank index should always be $1$. This makes it possible to interpret each element of decomposed weight as probabilities whether the nodes are in the specific group. Moreover, to ensure positiveness throughout the optimization process, a multiplication rule is used in backprop.

\begin{equation}
    W' = W \odot \frac{\nabla^-_{i j} - \lambda_i }{\nabla^+_{ij}}
\end{equation}

This method is useful since the decomposed weight is always positive and in low-rank form. After obtaining the pooling operator, the next step is obvious. We interpret low-rank weight as a pooling operator and transform nodes, edges and adjacencies.

\begin{eqnarray}
    S &\equiv& W\\
    \tilde{F}_{\textrm{(node)}} &=& S F_{\textrm{(node)}}\\
    \tilde{F}_{\textrm{(edge)}} &=& S F_{\textrm{(edge)}} S^T\\
    \tilde{A} &=& S A S^T
\end{eqnarray}

Yet there are some issues. One is that the optimization process is not accurate sometimes and does not reconstruct the original grouping matrix within a small margin. Another is that it costs huge amounts of computation resources that drags the whole learning process significantly.


\subsection{Specific Hyperparameters} \label{app_param}
In this subsection we introduce a set of hyperparameters used in experiments, more specifically. The parameters are expressed in the following tables. Table \ref{dataset_oriented} is dataset oriented parameters and table \ref{model_oriented} is model oriented parameters. Pooling ratios and hops are selected best performing parameter set after trying multiple set of combinations.
\begin{table}
  \caption{Hyperparameters on Dataset}
  \label{dataset_oriented}
  \centering
  \begin{tabular}{l|ll}
    \toprule
    Dataset    &    lr & Loss\\
    \midrule
    PLQY & 0.0001    & Mean Squared Error    \\
    $\lambda_{max}$  Films     & 0.00001    & Mean Squared Error\\
    $\lambda_{max}$  Solvents & 0.00001    & Mean Squared Error\\
    pIC50 & 0.00005        & Mean Squared Error\\
    Tox21 & 0.0001        & Binary Cross Entropy\\
    \bottomrule
  \end{tabular}
\end{table}

\subsection{Detailed Grouping Analysis}
\begin{figure*}[!htb]
    \centering
    \includegraphics[height=160mm, width=130mm, width=1\linewidth]{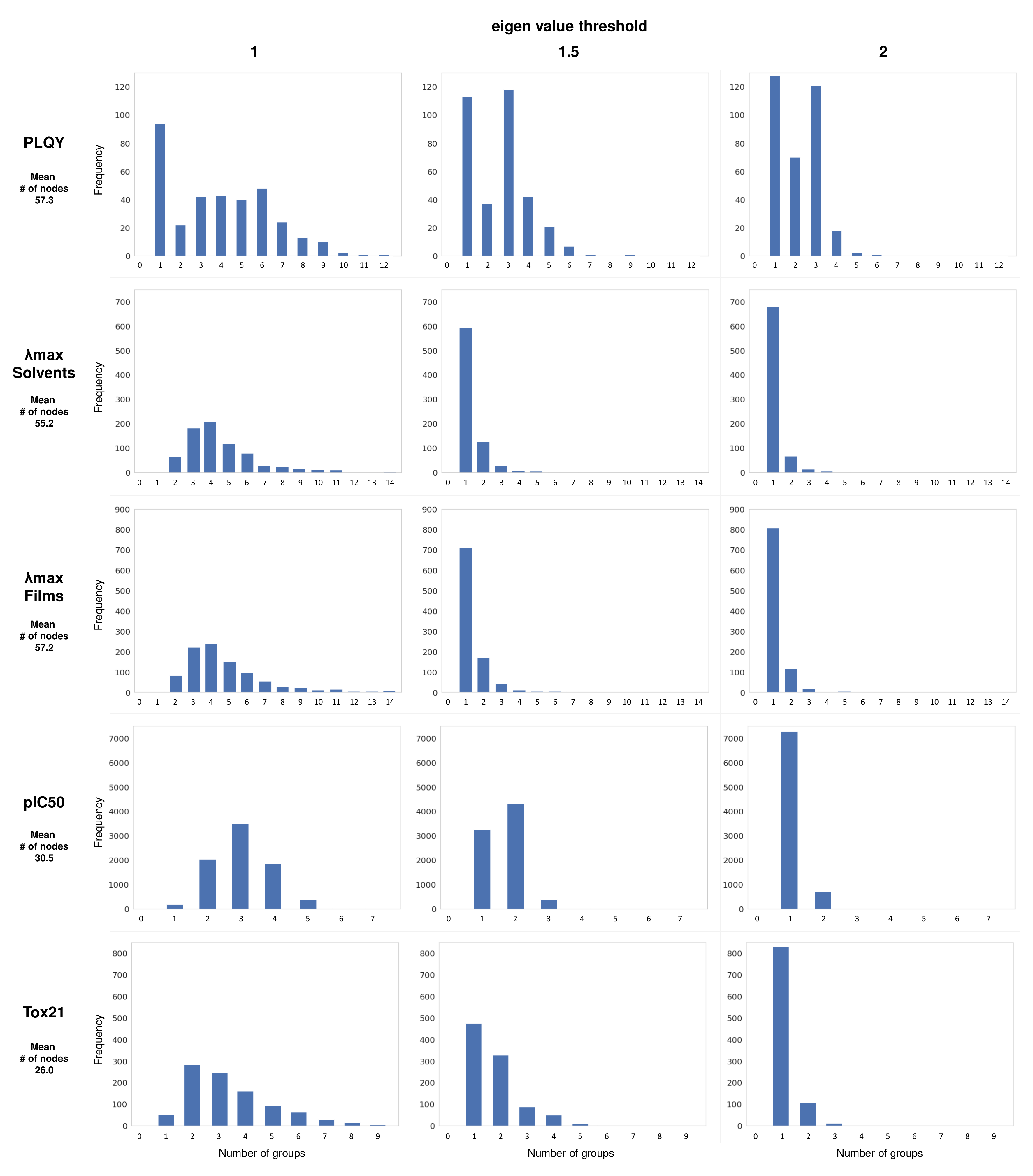}
    \caption{Histograms of effective number of clusters. The X-axis indicates the number of eigenvalues that exceed the threshold above, and Y-axis indicates the corresponding frequency.}
    \label{histogram}
\end{figure*}

In Figure~\ref{histogram}, we show histogram statistics of effective number of clusters obtained by imposing different thresholds on eigenvalues of the grouping matrix $M$. Note that an eigenvalue that equals 1 is equivalent to a cluster with 1 node. Thus, the number of eigenvalues of $M$ that exceed 1 indicates the number of clusters with an effective size larger than 1. We find that the number of effective clusters is significantly less than the average size of the molecules across all datasets, indicating that the model learns to effectively pool nodes and coarsens the molecular graph towards better performance.
\begin{gather}
    S \equiv O \sqrt{\Lambda}\\
    D = S^T S = \sqrt{\Lambda} O^T  O \sqrt{\Lambda} = \Lambda
\end{gather}

\end{document}